\documentclass[letter, 10pt, conference]{ieeeconf}      % Use this line for a4 paper

\IEEEoverridecommandlockouts                              % This command is only needed if 
\makeatletter
\let\NAT@parse\undefined
\makeatother

\usepackage{cite}
\usepackage{graphics} % for pdf, bitmapped graphics files
\usepackage{epsfig} % for postscript graphics files
\usepackage{mathptmx} % assumes new font selection scheme installed
\usepackage{times} % assumes new font selection scheme installed
\usepackage{amsmath} % assumes amsmath package installed
\usepackage{amssymb}  % assumes amsmath package installed
\usepackage{colortbl}
\usepackage{xcolor}
\usepackage{hyperref}

\title{\LARGE \bf
ROADFIRST: A Comprehensive Enhancement of the Systemic Approach to Safety for Improved Risk Factor Identification and Evaluation
}

\author{Shriyan Reyya$^{1}$ and Yao Cheng$^{2}$% <-this % stops a space
\thanks{$^{1}$Shriyan Reyya is with the Department of Civil and Environmental Engineering, University of Maryland, College Park and Montgomery Blair High School, Silver Spring, MD
        {\tt\small shriyanr@umd.edu}}%
\thanks{$^{2}$Yao Cheng Ph.D. is with the Department of Civil and Environmental Engineering, University of Maryland, College Park
        {\tt\small ycheng09@umd.edu}}%
}

\begin{document}

\maketitle
\thispagestyle{empty}
\pagestyle{empty}

%%%%%%%%%%%%%%%%%%%%%%%%%%%%%%%%%%%%%%%%%%%%%%%%%%%%%%%%%%%%%%%%%%%%%%%%%%%%%%%%
\begin{abstract}

Many agencies have adopted the FHWA-recommended systemic approach to traffic safety, an essential supplement to the traditional hotspot crash analysis which develops region-wide safety projects based on identified risk factors. However, this approach narrows analysis to specific crash and facility types. This specification causes inefficient use of crash and inventory data as well as non-comprehensive risk evaluation and countermeasure selection for each location. To improve the comprehensiveness of the systemic approach to safety, we develop an enhanced process, ROADFIRST, that allows users to identify potential crash types and contributing factors at any location. As the knowledge base for such a process, crash types and contributing factors are analyzed with respect to features of interest, including both dynamic and static traffic-related features, using Random Forest and analyzed with the SHapley Additive exPlanations (SHAP) analysis. We identify and rank features impacting the likelihood of three sample contributing factors, namely alcohol-impaired driving, distracted driving, and speeding, according to crash and road inventory data from North Carolina, and quantify state-wide road segment risk for each contributing factor. The introduced models and methods serve as a sample for the further development of ROADFIRST by state and local agencies, which benefits the planning of more comprehensive region-wide safety improvement projects.

\end{abstract}

%%%%%%%%%%%%%%%%%%%%%%%%%%%%%%%%%%%%%%%%%%%%%%%%%%%%%%%%%%%%%%%%%%%%%%%%%%%%%%%%
\section{INTRODUCTION}
The traditional hotspot crash analysis is a prominent method in traffic safety, relying on the site-specific analysis of high-crash-density locations to identify root causes and implement countermeasures. To proactively develop safety improvement projects for road segments with high crash risks before losses occur, the Federal Highway Administration (FHWA) recently recommended the systemic approach, which acknowledges that even sparse crash records within a wider region can provide valuable insights if properly analyzed \cite{reliability}, \cite{NAP26032}.

\begin{figure}[htbp]
\centerline{\includegraphics[width=\linewidth]{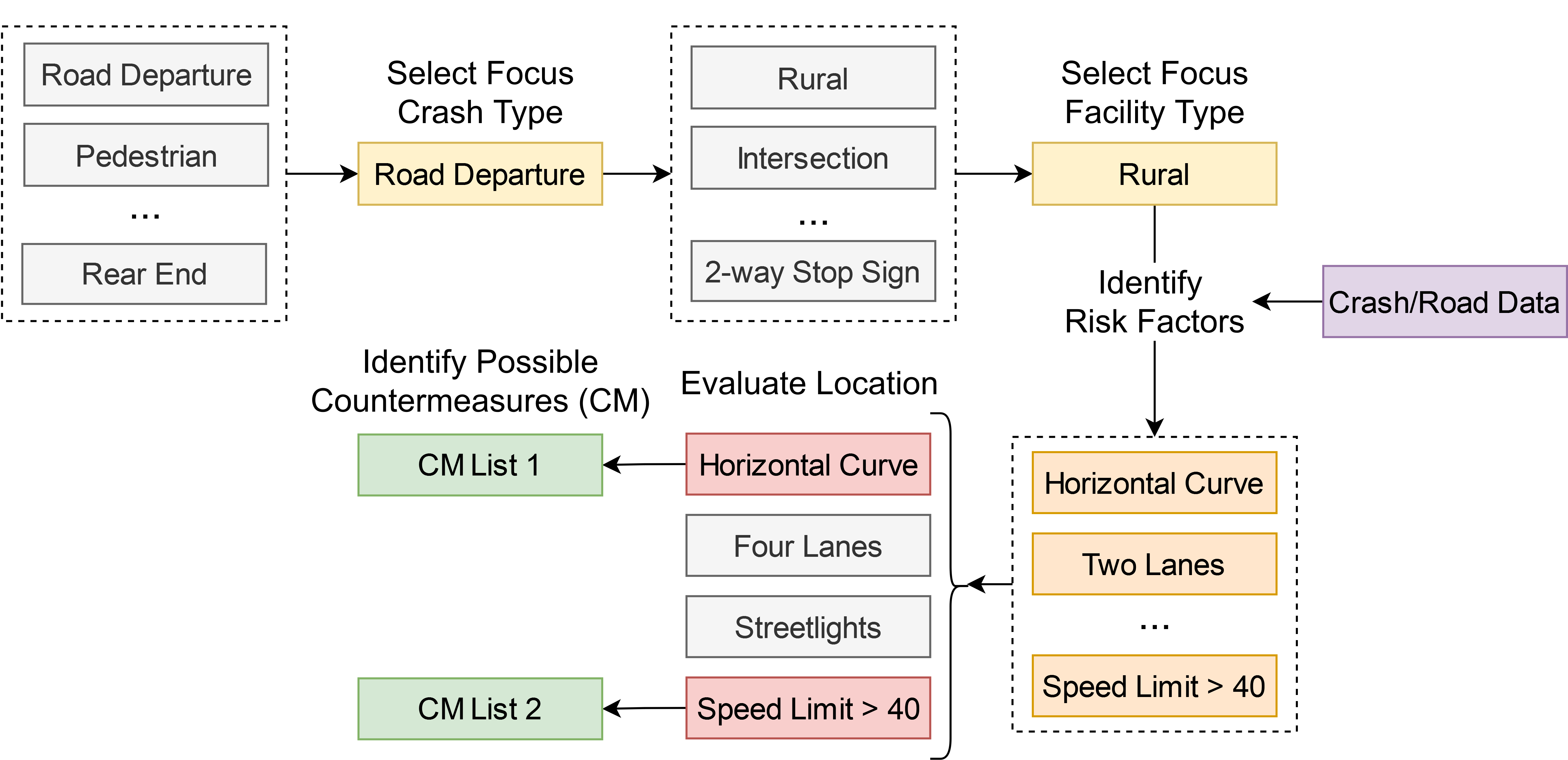}}
\caption{Systemic safety implementation process.}
\label{fig:systemic}
\end{figure}

As shown in Figure \ref{fig:systemic}, the systemic approach first requires determining specific focus crash and facility types. Next, generally with help from statistical analysis of crash and road inventory data, risk factors (physical road features characterized to indicate risk) are identified. Safety improvement projects and countermeasures are identified and developed based on examining regional road inventory and proactively selecting locations containing identified risk factors, regardless of crash frequency \cite{selectiontool}.

Such a process, despite its effectiveness in proactively developing region-wide safety improvement projects, presents the following limitations:
\begin{itemize}
    \item By limiting focus crash and facility types, each analysis usually takes advantage of a limited fraction of the entire crash dataset at a time, not harnessing the availability of extensive crash records.
    \item The process may fail in identifying common risk factors that are not sufficiently outstanding for specific crash or facility types but are constantly significant across types.
    \item At a given location, the process assesses the risk for a crash type that has been pre-specified from the region-wide analysis, but is not necessarily dominant at that specific subject location, causing the risks of other crash types to be overlooked.
    \item Since most identified risk factors are related to geometric or traffic conditions of specific locations \cite{selectiontool}, the process excels in identifying countermeasures that target road features while hardly selecting those targeting other nonphysical factors contributing to crashes, (e.g., alcohol-impaired driving, speeding, etc.). For example, warning signs for texting and driving, which do not directly address a road feature, may be overlooked in the selected countermeasure list.
\end{itemize}

\begin{figure}[htbp]
\centerline{\includegraphics[width=\linewidth]{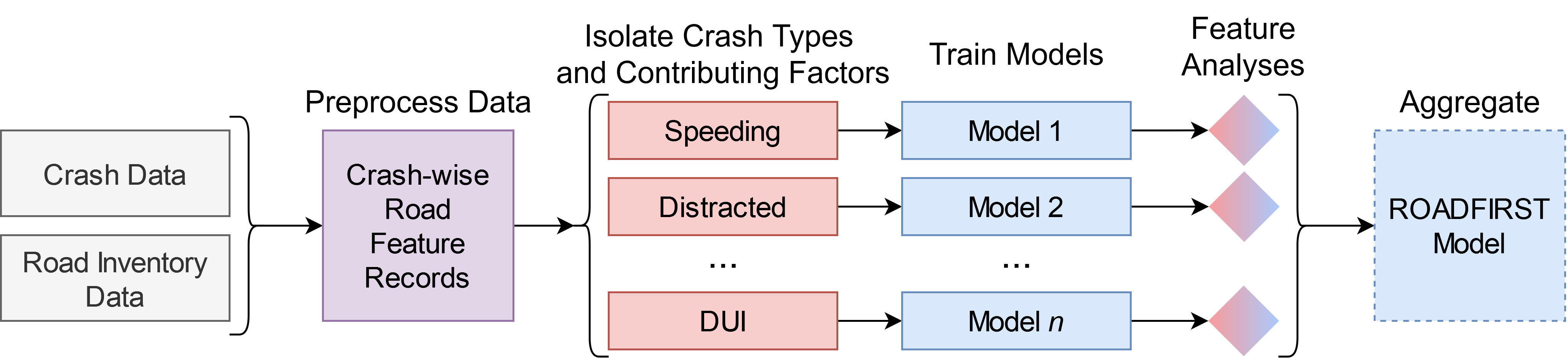}}
\caption{ROADFIRST development and analysis.}
\label{fig:modeldev}
\end{figure}

\begin{figure}[htbp]
\centerline{\includegraphics[width=\linewidth]{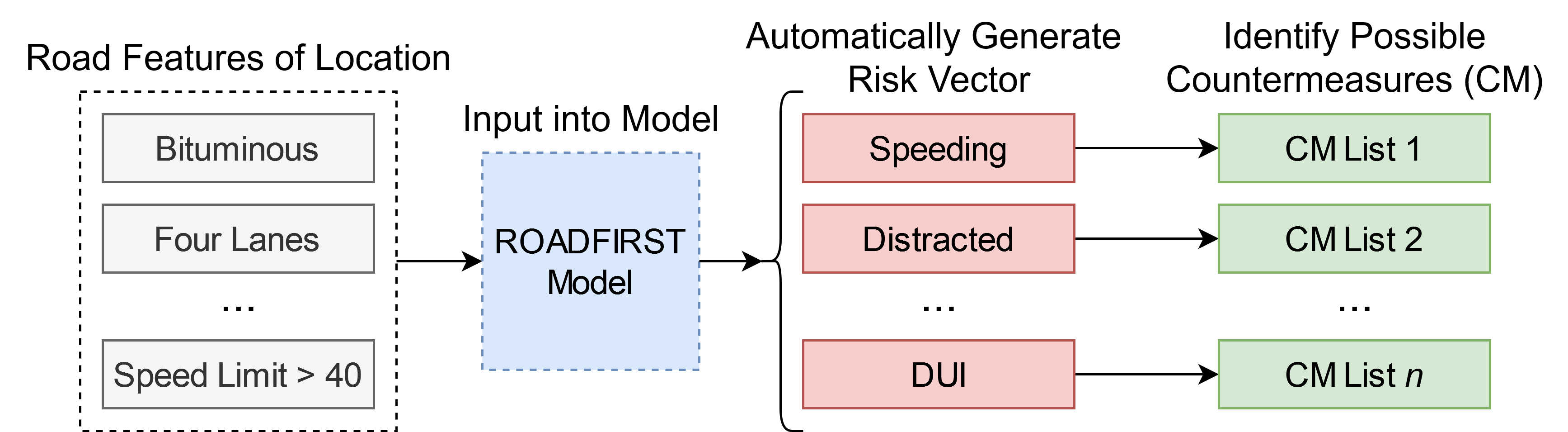}}
\caption{ROADFIRST implementation.}
\label{fig:roadfirst}
\end{figure}

To address the above concerns, we propose ROADFIRST, an enhancement to the state-of-the-practice systemic approach that takes a road-first approach to risk evaluation, as shown in Figures \ref{fig:modeldev} and \ref{fig:roadfirst}. Via the ROADFIRST methodology, agencies develop a region-specific model to analyze both crash types and contributing factors potentially targeted by a wide range of countermeasures, emphasizing prevalent nonphysical factors such as alcohol-impaired driving and speeding. As detailed in Figure \ref{fig:modeldev}, influential features for each contributing factor are analyzed, providing an overview of key region-wide risk factors, including both dynamic crash-specific features and static road features.

Notably, the results from the developed prediction models can be used to evaluate the risk of individual locations with respect to various types of crashes, as shown in Figure \ref{fig:roadfirst}. From this point, countermeasures can be developed with every risk in mind. The output from such a process serves not only as a reliable resource for transportation agencies to identify region-wide safety improvement projects with accurately targeted countermeasures, but also as a convenient tool to evaluate the safety risk at specific locations.

\section{LITERATURE REVIEW}
The FHWA maintains records of exemplary systemic safety practices, one of which was published in a report describing the New York State DOT approach to mitigating lane departure crashes on state highway systems \cite{nydotlane}. Using a proportional analysis of accident counts, they identified three major risk factors. Countermeasures were then selected with the help of the National Cooperative Highway Research Program 500 Guides.

The FHWA has additionally published several reports of systemic safety implementations, many of which employ the idea of analyzing risk based on road features in lieu of the end-to-end process. Illinois developed a Safety Performance Function (SPF) for systemic analysis of risk and countermeasure evaluation as reported in \cite{illinoisnoteworthy}. As documented in  \cite{kentuckycabinet}, Kentucky consulted the Systemic Safety Project Selection Tool, a comprehensive guide to the conventional implementation of systemic safety, to inform their process, as initial attempts were unsuccessful. Sourcing data from various agencies across five counties, they identified risk factors for roadway departure crashes on horizontal curves on rural county roads and then conducted an in-depth analysis to quantify the risk of their roads. In \cite{assessmentlocal}, assessments of Louisiana, Nebraska, and Ohio all detailed establishing Local Technical Assistance Programs to administer local road infrastructure safety projects using the systemic process. Missouri used a data-driven approach to evaluate different lane painting methods, as reported in \cite{missouricase}. They tested two separate methods on rural roads and analyzed crashes over two years, concluding that one method had a statistically significant improvement while the other did not.

Some other case studies describe full, end-to-end processes. Kansas targeted collisions with fixed objects as their focus crash type and used a proportional analysis to evaluate locations. They sourced data from local agencies and selected countermeasures with an emphasis on low costs \cite{kansas}. Minnesota consulted the Systemic Safety Project Selection Tool to inform their statewide, comprehensive crash analysis and used the results of this analysis to focus on county systems and rural crashes. They consulted literature to identify risk factors for each of curves, intersections, and segments. Countermeasures were screened using crash data, effectiveness, cost, and agency policies \cite{minnesotacase}. The Public Works Department of Thurston County, Washington, reviewed crash data maintained by the County Advisory Board, the Washington State DOT, aerial photography, and geographic information system files. From this review, they identified their focus crash and facility types to be roadway departure crashes on horizontal curves and signed curves on arterials and collectors, respectively. They consulted the Systemic Safety Project Selection Tool to identify risk factors and countermeasures and created a scoring mechanism for quantifying risk along roadways \cite{thurstoncounty}. 

Apart from case studies, independent state sponsored reviews and implementations are reported as well. Authors in \cite{ncodt} reviewed several practices of New York, Minnesota, Ohio, Michigan, and Florida, which all used data-driven methods to identify risk factors with respect to accidents. In order to identify risk factors for their own implementations, they compiled several databases, including crash data, road inventory data, pedestrian infrastructure, pedestrian exposure, demographics, land use, and vehicle speed. After reviewing the scope of the data, they decided to focus only on urban roadway segments. For statistical modeling, they developed two separate methods to accomplish two different tasks. The first was to quantify the effect of each individual factor at intersections, and the second was to estimate the pedestrian exposure at each intersection. Both used negative binomial regression as suggested by literature. Employing this methodology, they were able to make a comprehensive list of risk factors.

As a part of Montgomery County's Vision Zero mission, authors in \cite{moco} implemented a systemic predictive safety analysis of pedestrian crashes. They produced two SPFs for two different facility types and used negative binomial regression on police-recorded crash data and county network data, identifying key transportation-related exposure variables. Similarly, the Massachusetts Department of Transportation conducted systemic safety analysis for midblock pedestrian crashes, incorporating crash, roadway, transit, census, and equity data. They identified three focus facility types and used binomial logit regression model to determine risk factors \cite{massdot}.

The FHWA also presents a general implementation methodology with use of National Oceanic and Atmospheric Administration and US Census Bureau data. They suggest the use of negative binomial regression, random forest, or classification and regression trees to identify risk factors directly based on accidents of the focus crash type \cite{contribfocus}.

In summary, the systemic approach to safety has been widely applied in different states under various levels of data availability constraints and with a variety of modeling techniques. However, most implementations are focused on infrastructure and traffic-related risk factors, overlooking nonphysical crash contributing factors. This tendency may undesirably prevent some effective countermeasures, especially those involving enforcement and education, from being selected. Moreover, most studies focus on a specific crash and facility type, limiting the identification of risk factors that potentially demonstrate significant impacts across the network. Therefore, this study aims to develop a procedure that can fully leverage the available data to build a reliable relation between physical road features and various categories of crash contributing factors, enhancing the applicability across the entire network and comprehensiveness of the considered countermeasures.

\section{ROADFIRST}
ROADFIRST works to provide a more comprehensive region-wide risk factor identification by supplementing analysis focuses and including all available crash data as well as road inventory data. Targeting the most prevalent crash and facility types, as outlined in the systemic approach to safety, is certainly a priority. However, the effort to evaluate the risk of any particular road should not be limited to just the area-wide prevailing crash type, as a greater danger may exist, one not prevalent across the region, that will be completely missed. Incorporating a more comprehensive risk evaluation allows for a risk profile to be built and stored. In other words, if agencies are analyzing a road for a specific crash risk, why not also check if it presents risks for other crash types, regardless of how common these types are? Such a comprehensive analysis justifies the need for ROADFIRST.

Moreover, ROADFIRST includes nonphysical contributing factors as targets in addition to accident types, allowing for a much easier pipeline to countermeasures, many of which are based on such factors in lieu of accident types. For example, in a certain region, speeding crashes may be comprised of road departures, rollovers, and pedestrian crashes. Suppose that of these three crash types, road departure is the only prevalent type in said region. Now, two main issues arise. Firstly, locations at risk of speeding crashes but not road departures will be ignored in analyses. Secondly, locations at risk of road departures as a result of speeding will not be improved optimally, as speeding-targeted countermeasures may not be considered whatsoever. Including speeding as a focus from the very beginning mitigates these issues and presents the opportunity to conduct more comprehensive improvements. This reasoning can be reversed as well, and multiple prevalent crash types that may represent a nonphysical contributing factor can be collapsed into that factor to ease countermeasure identification, as many countermeasures are based specifically on nonphysical factors.

Finally, ROADFIRST recommends including dynamic features as predictors and using more advanced analysis methods than those currently in practice. Day-to-day resource allocation and enforcement distribution can be heavily optimized by including these dynamic conditions into analysis. As for the analysis itself, tree-based models and independent interpretation metrics are both more complex than many current state-of-the-practice methods, many of which are simple proportional observations. They are not as simple to implement, but do not incur any additional cost other than computational. The payoff might be incredibly valuable considering the improved effectiveness of the project development.

\section{DATA AND METHODS}
To demonstrate the potential of the ROADFIRST methodology, we developed prototypes based on three key contributing factors: alcohol-impaired driving, distracted driving, and speeding. 

For each contributing factor, two models were trained. The purpose of the first model for each factor is to conduct analyses of overall most influential features, allowing ROADFIRST to be used to inform enforcement decisions (e.g. where and when to watch for alcohol-impaired driving). Therefore, this model format was trained with both static road features and dynamic crash-specific features as regressors. Subsequently, this type of model will be referred to as the combined feature model. The purpose of the second model, which can also be used for feature analysis on an agency basis, is automated risk evaluation. The training format only includes static road features as regressors and will be referred to as the road feature model.

\subsection{Data Preprocessing} Crash and road inventory data were sourced from the Highway Safety Information System (HSIS) Database, which is collected and maintained by the FHWA. This database contains detailed crash records and road inventory data from 9 states in the United States. Each crash record has a unique identification that can be matched to a specific road segment in the inventory data, allowing crashes to be analyzed by the physical road features present. Data maintained by North Carolina contain the most information and were therefore used in training models for each contributing factor. While some findings are region specific, others are more generalized, so data maintained by Maine and Minnesota, which are each partially overlapping with those of North Carolina in terms of features included, were used for validation.

The relevant data in each state consist of four files per year: the crash file, unit file, person file, and roadway file, each of which contains information pertaining to the crash, unit, people, and roadway, respectively.  

\begin{figure}[htbp]
\centerline{\includegraphics[scale = 0.07]{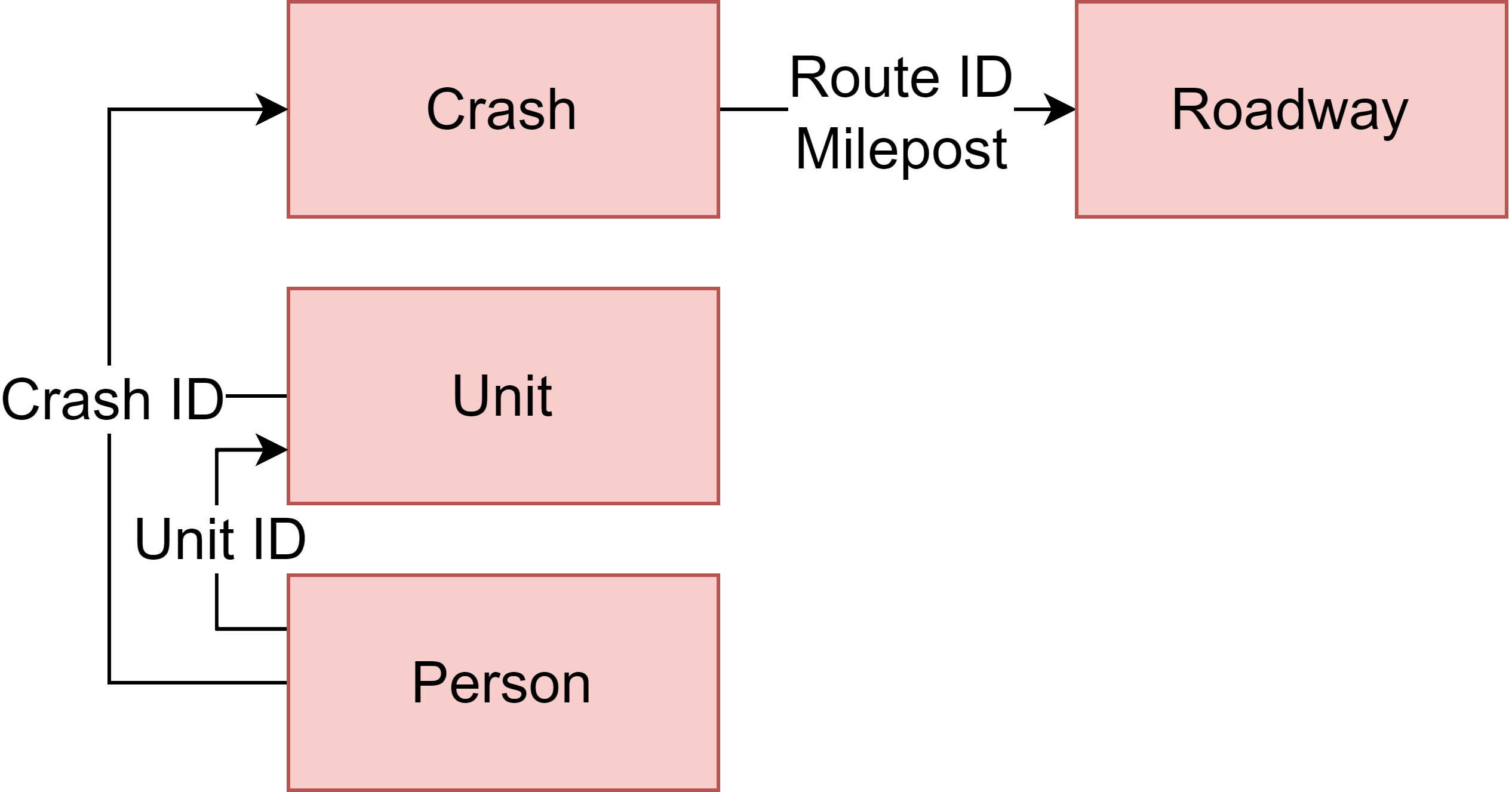}}
\caption{Data files and connections in HSIS Database.}
\label{fig:connections}
\end{figure}

As shown in Figure \ref{fig:connections}, the data in each of the crash, unit, and person files are connected by identifiers, and relevant features from each were combined into one file. Road inventory features were incorporated by mapping the entries in the roadway file using route identifications and milepost numbers. All available road features were included, as well as dynamic features independent of driver or vehicle information describing data before or at the time of crash, such as weather and lighting condition.

In all of the states considered, a significant proportion of road features originate in the crash and unit files rather than road inventory data. Due to the dynamic nature of these files, any temporary road feature changes existing at the time of the crash (e.g. surface type, number of lanes, etc.) are reflected. We accordingly chose to structure the modeling dataset in the format where each row represents a crash, in contrast to the format where each row represents a road segment. An added advantage of this format is that the complex multi-variate patterns in and between dynamic features and road features are captured more easily than would be possible with aggregating by segment.

Note that modeling with a crash-wise dataset affects subsequent interpretation by introducing the mandatory condition of a crash occurring. Namely, given that a crash has occurred, a model allows for the identification and understanding of the specific features associated with the crash being the result of a particular contributing factor.

Categorical features were split up into binary dummy variables, allowing for simplified interpretation in later stages of the process. To address possible multicollinearity, especially in dummy variables and between similar road features recorded in separate files, we used Variance Inflation Factor (VIF) values. The VIF value $\text{VIF}_{i}$ of the $i$th independent variable in a dataset is derived by training a linear regression model and analyzing variance, specifically 

\begin{equation}
    {\text{VIF}_{i} = \frac{1}{1 - R^{2}_{i}}},
\end{equation} 

where $R^{2}_{i}$ represents the proportion of variance in the $i$th independent variable that is associated with the other independent variables in the model. $R^{2}_{i}$ is a good indicator of multicollinearity; generally, the greater $\text{VIF}_{i}$ is, the more severe the multicollinearity for the $i$th independent variable. 

With the goal of retaining as many features as possible to keep later interpretation stages easy, we did not use common cutoff thresholds for VIF values. Instead, following guidelines discussed in \cite{vif}, a holistic approach was implemented. Referencing the dataset correlation matrix, individual pairs of columns with high correlations were reviewed, and either the variables were combined, or the one with lower interpretive importance (a subjective, agency-dependent decision) was removed. Variables with VIF values greater than 10 that correlated with each other, although not as severely, were identified and addressed in the same manner. Additionally, if several dummies derived from the same variable had a high VIF value, only the one with the highest value was removed. The dataset VIF values were recalculated and this process was iterated until all VIF values were below 10. This approach allows for high VIF values to lower over each iteration while severe correlations are addressed, preventing a more widespread removal of columns that might have resulted from using a cutoff value. 

\subsection{Descriptions of Preprocessed Data}
The preprocessed North Carolina dataset, used for model development, consists of 202,822 crashes and 93 variables, including route identifications, mileposts, and a binary target variable for each crash contributing factor, i.e., alcohol-impaired driving, distracted driving, and speeding. The 88 independent variables were derived from 25 variables in the original HSIS database, with 80 of the independent variables being binary dummies.

The verification datasets are derived from two other states and each has significantly fewer crashes and variables. Minnesota’s dataset contains 48,907 crashes from 2020 and 58 variables, while Maine’s dataset contains 12,258 crashes from 2021 and 44 variables. Each dataset includes route identifications, mileposts, and one binary target variable: alcohol-impaired driving in Minnesota and speeding in Maine. No reliable distracted driving feature was available for verification.

Before training each respective model, the route identification, milepost, and non-target crash contributing factor variables were dropped. For the road feature models, all dynamic features were dropped. 

\subsection{Modeling and Interpretation} Similar studies have been conducted in the context of understanding crash contributing factors. The FHWA notes that tree based algorithms, such as decision trees and random forests, are able efficiently identify complex interactions between variables \cite{contribfocus}. This idea is supported by \cite{AHMED2023100814}, where multiple algorithms were tested to identify risk features of New Zealand accidents, and random forest was found to be the most accurate. Moreover, random forest is considered optimal for analysis of large datasets with numerous independent variables, which is especially useful in the context of understanding traffic risk \cite{info11050270}, \cite{su14031729}. Therefore, we adopted the random forest algorithm to identify relationships between contributing factors and relevant features and evaluate location risks.

In all crash-contributing factors considered, there was a data imbalance between crashes caused by the respective factor and those not caused by it. To address this imbalance, we implemented a combination of Random Undersampling the Majority Class (RUMC) and the Synthetic Minority Oversampling Technique (SMOTE). Both methods have been used individually in traffic safety studies \cite{AHMED2023100814}, \cite{infrastructures5070061}. The original authors of SMOTE recommend combining these techniques to prevent overfitting and achieve optimal results \cite{Chawla_2002}. Additionally, traditional SMOTE generates continuous values but not categorical ones. To address this, we used SMOTE-NC, a modified algorithm that accounts for both continuous numerical and categorical variables \cite{smotenc}. The data were split into an 80\% training set and a 20\% testing set.

Specific positive and negative feature influences in random forest, and many other models, are hidden, but one solution to understanding these influences is SHapley Additive exPlanations, or SHAP. SHAP relies on the game theory-based Shapley values, which are numerical representations of contributions of each feature to a prediction made by the model. SHAP requires re-training a given model on all subsets $S \subseteq F$, where $F$ is the set of all features. Feature-wise models are trained such that every $S$ is trained with a given feature present and with that same feature withheld. This process, when iterated over all subsets $S$ of $F$, provides an overview of each feature's contribution to each prediction made by the model. The Shapley value $\Phi_{i}$ of the $i$th feature is therefore calculated as

\begin{equation}
    {\Phi_{i} = \Sigma_{S\subseteq F\setminus \{i\}}\frac{|S|!(|F| - |S| - 1)!}{|F|!}[f(S \cup \{i\}) - f(S)]}
\end{equation} where $f$ is the respective model prediction under each iteration \cite{shap}.

SHAP's results are preferable to other explainable machine learning (XML) methods, as demonstrated in \cite{AHMED2023100814}, where methods including SHAP, Local Interpretable Model-agnostic Explanation, and Layer-wise Relevance Propagation were compared. They note that SHAP's accuracy, consistency, and visualization capabilities make it the best option for XML in the context of crash-contributing factors.

\section{RESULTS}

The results of both the SHAP analyses and the state-wide risk evaluations are presented. The SHAP analyses are derived from the combined feature models and the evaluative results from the road feature models. 

\subsection{Risk Factor Identification with SHAP Analysis}
The SHAP analysis results for each of the three non-physical crash-contributing factors are shown and interpreted in plots of influential features and their Shapley values across every prediction. Positive Shapley values indicate a positive model influence, and the numerical magnitude is represented in a color gradient scale. For binary features, there is no gradient; red represents the presence of the respective feature at the time of crash and blue represents the lack of it.

The major findings from the analysis with respect to the three crash contributing factors are presented below.

\subsubsection{Alcohol-Impaired Driving} 
\begin{figure}[!htbp]
% \centerline{\includegraphics[scale = 0.4]{figs/general_alc_named_summary_abridged_no_time.png}}
\centerline{\includegraphics[scale = 0.4]{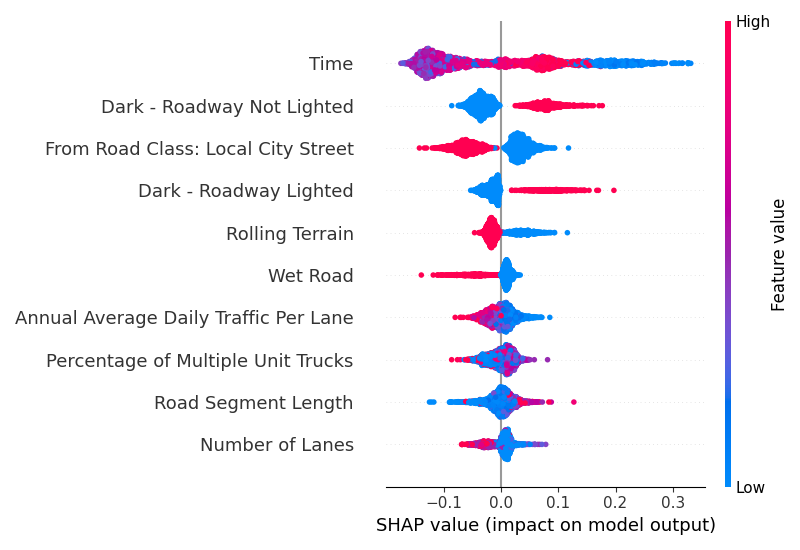}}
\caption{North Carolina alcohol-impaired driving model SHAP summary.}
\label{fig:duisum}
\end{figure}

\begin{figure}[!htbp]
% \centerline{\includegraphics[scale = 0.4]{figs/alc_summary_named_abridged_notime.png}}
\centerline{\includegraphics[scale = 0.4]{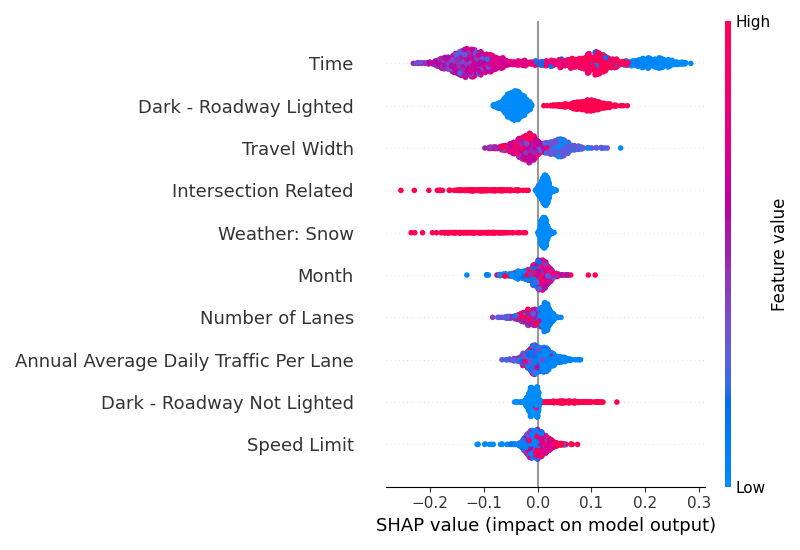}}
\caption{Minnesota alcohol-impaired driving model SHAP summary.}
\label{fig:duisummin}
\end{figure}

\begin{figure}[!htbp]
\centerline{\includegraphics[scale = 0.4]{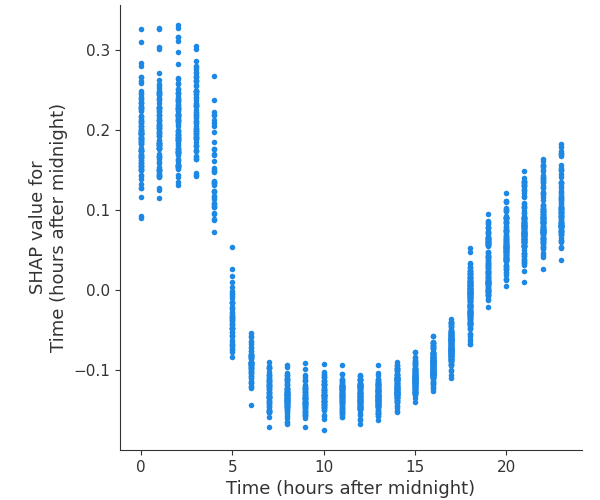}}
\caption{Comparison of the Time feature and its respective Shapley values in the North Carolina model.}
\label{fig:duitime}
\end{figure}

As shown in Fig. \ref{fig:duisum} and Fig. \ref{fig:duisummin}, a consistent and significant influential feature is the time of day. In Fig. \ref{fig:duitime}, it is clearly shown that crashes occurring between roughly 11:00 PM and 4:00 AM are substantially more likely to be classified as caused by alcohol-impaired driving, whereas the opposite effect is observed for hours between 5:00 AM and 10:00 PM. Similarly, crashes occurring in the dark, regardless of roadway lighting, are more likely to be classified as caused by alcohol-impaired driving. Statistics from the National Highway Traffic Safety Administration (NHTSA) support this finding, stating that fatal alcohol impairment rates were 2.8 times higher at night than during the day \cite{nhtsalc}. The precedence of time of day over lighting conditions may be attributed to societal norms regarding drinking habits throughout the year, irrespective of seasonal lighting changes.

The presence of a wet road significantly decreases likelihood of being classified as caused by alcohol-impaired driving, which is reflected in the Minnesota model with the equivalent snowy weather feature. 

This finding is supported by NHTSA statistics as well, which show that 91\% of alcohol-impaired driving crashes happened in clear or cloudy weather in 2021 \cite{nhtsalc}. 

Annual average daily traffic per lane, a measure of traffic density, has a negative correlation with the likelihood of a crash being classified as caused by alcohol-impaired driving. Similarly, the number of lanes in both the North Carolina and Minnesota models shows the same effect.

Authors in \cite{KEALL2005816} found that high-volume roads are least affected by alcohol-impaired driving crashes, corroborating this result.

Crashes occurring on roads connected to local city streets are unlikely to be classified as caused by alcohol-impaired driving. In contrast, crashes occurring on roads not connected to local city streets demonstrate the opposite effect.

\subsubsection{Distracted Driving} 
\begin{figure}[htbp]
\centerline{\includegraphics[scale = 0.4]{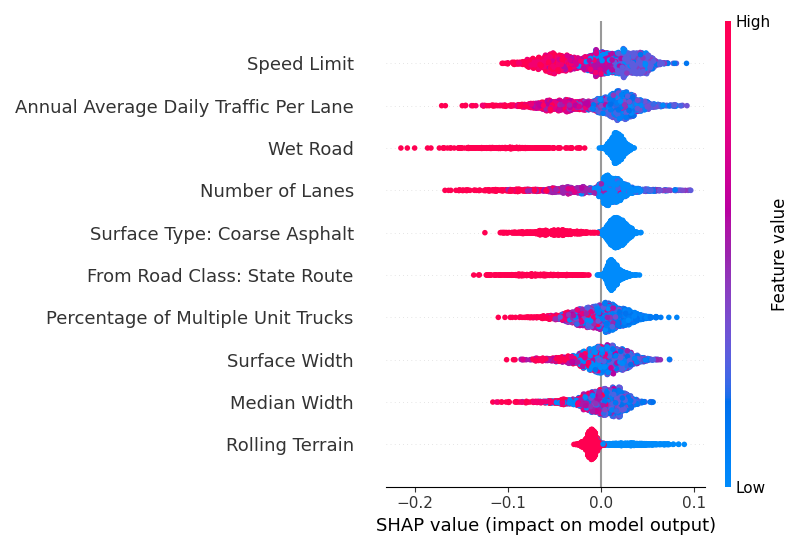}}
\caption{North Carolina distracted driving model SHAP summary.}
\label{fig:distracted}
\end{figure}

\begin{figure}[htbp]
\centerline{\includegraphics[scale = 0.4]{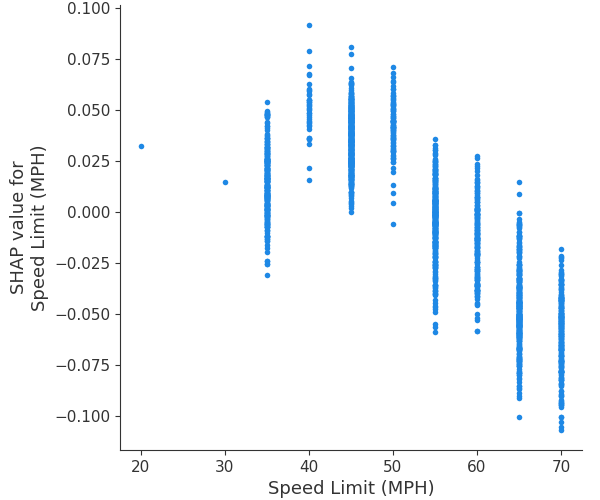}}
\caption{Comparison of the Speed Limit feature and its respective Shapley values in the North Carolina distracted driving model.}
\label{fig:distracted_speed}
\end{figure}

As depicted in Fig. \ref{fig:distracted}, crashes occurring on roads with high speed limits are very unlikely to be attributed to distracted driving. The effects of low and mid-range speed limits are more ambiguous, but generally follow the overall negative correlation with speed limit and distracted driving likelihood. Fig. \ref{fig:distracted_speed} suggests that the maximum risk is at speed limits between 40 and 50 miles per hour. This risk slightly drops on roads with speed limits below 40 miles per hour, possibly because, regardless of distractions, crash risk is lowered at lower speeds.

Traffic density, number of lanes, percentage of multiple unit trucks, surface width, and median width all follow a similar pattern of negative correlation with likelihood of a particular crash being the result of distracted driving. The presences of a wet road and coarse asphalt follow this pattern as well, suggesting that, in general, the fewer features that deviate from ordinary exist on the road, the more distracted the driver tends to be.

\subsubsection{Speeding} 
\begin{figure}[!htbp]
\centerline{\includegraphics[scale = 0.4]{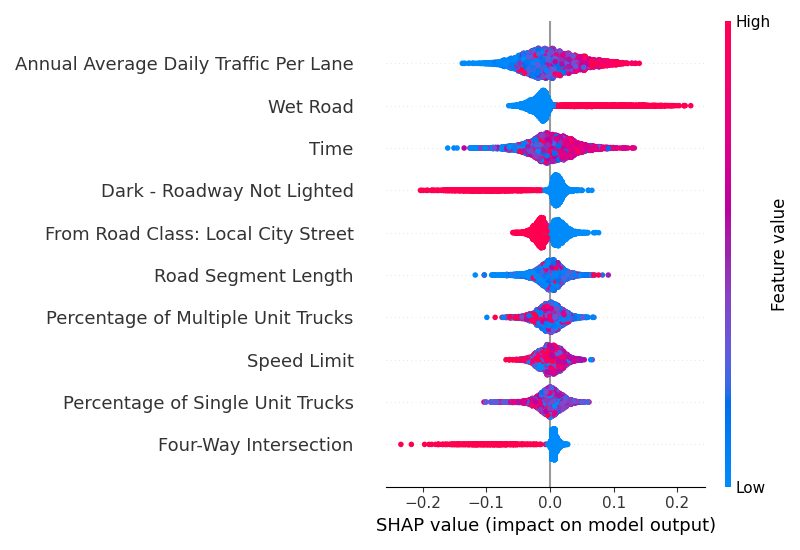}}
\caption{North Carolina speeding model SHAP summary.}
\label{fig:speedingnc}
\end{figure}

As shown in Fig. \ref{fig:speedingnc}, traffic density is very highly correlated with the likelihood of a crash being classified as caused by speeding, possibly caused by the higher number of objects to crash into when speeding. 

\begin{figure}[!htbp]
\centerline{\includegraphics[scale = 0.4]{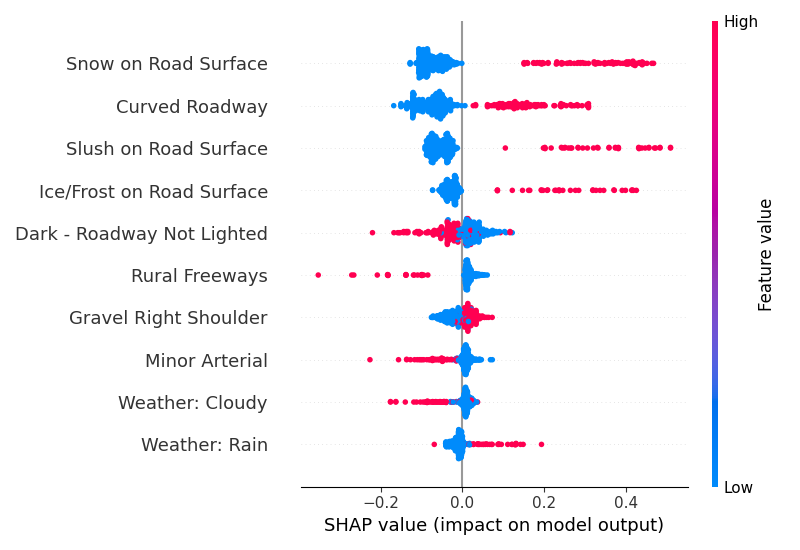}}
\caption{Maine speeding model SHAP summary.}
\label{fig:speedingme}
\end{figure}

Additionally, weather plays a large role, as shown in both the North Carolina model and the Maine model in Fig. \ref{fig:speedingme}. Crashes on wet roads or roads with snow, slush, ice, or frost highly positively influence the likelihood of being classified as caused by speeding, likely due to loss of control at such speeds in such conditions. 

These findings are supported by NHTSA statistics, stating that, in 2021, 22\% of fatal speeding crashes occurred on wet roads, 32\% when there was snow or slush on the road, and 33\% on roads with ice or frost \cite{nhtsaspeeding}.

Crashes occurring on dark, unlighted roadways significantly lower the likelihood of being classified as caused by speeding, possibly because drivers tend to avoid speeding in such conditions because of perceived uncertainty. Similarly, crashes at four-way intersections show the same effect, likely because there is insufficient space to accelerate properly before a crash, suggesting that other factors are more likely to contribute to these incidents.

Crashes occurring on roads coming from local city streets also lower the likelihood of being attributed to speeding, possibly because drivers tend to follow their previous driving behavior the most, which, on local city streets, would generally be low speeds.

\subsection{Road Feature Model Implementation}
The primary purpose of the road feature models is the automation of risk evaluation. Given an input of a road segment as a set of features, the models output a confidence percentage of how likely a crash occurring on that segment is to result from a particular crash contributing factor. 

\begin{figure}[!htbp]
\centerline{\includegraphics[width = \linewidth]{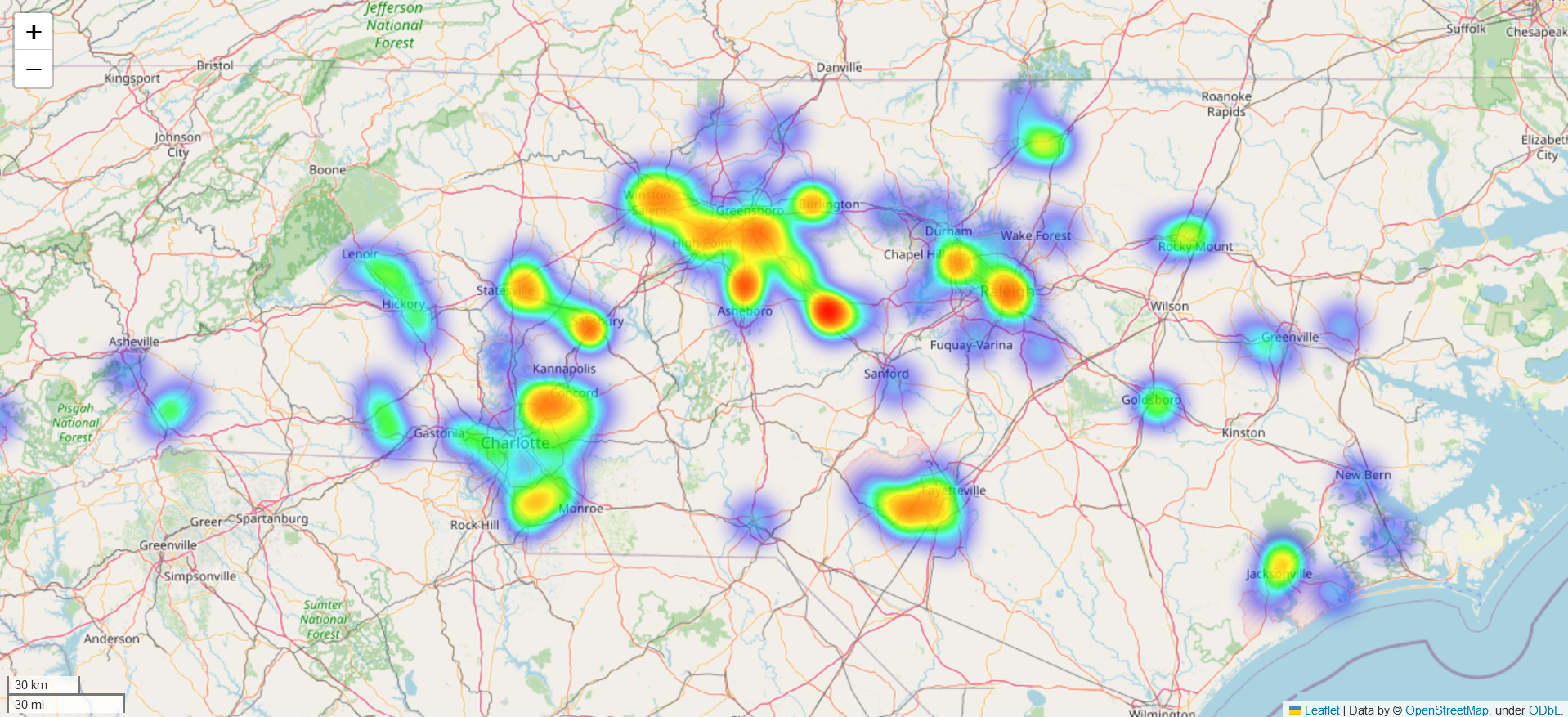}}
\caption{Spatial heat-map weighted by classification confidence of potential crashes being caused by alcohol-impaired driving.}
\label{fig:alcroad}
\end{figure}

\begin{figure}[!htbp]
\centerline{\includegraphics[width = \linewidth]{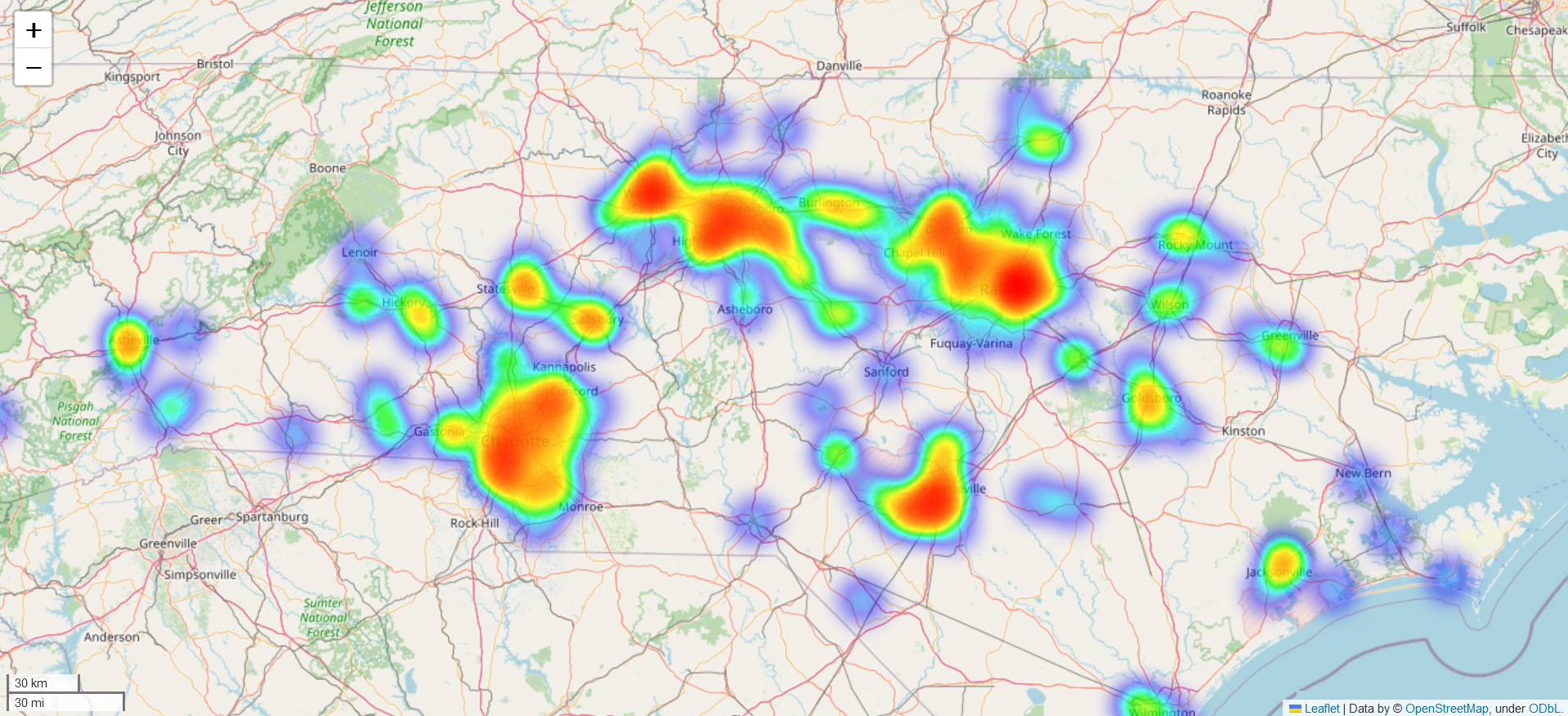}}
\caption{Spatial heat-map weighted by classification confidence of potential crashes being caused by distracted driving.}
\label{fig:distroad}
\end{figure}

\begin{figure}[!htbp]
\centerline{\includegraphics[width = \linewidth]{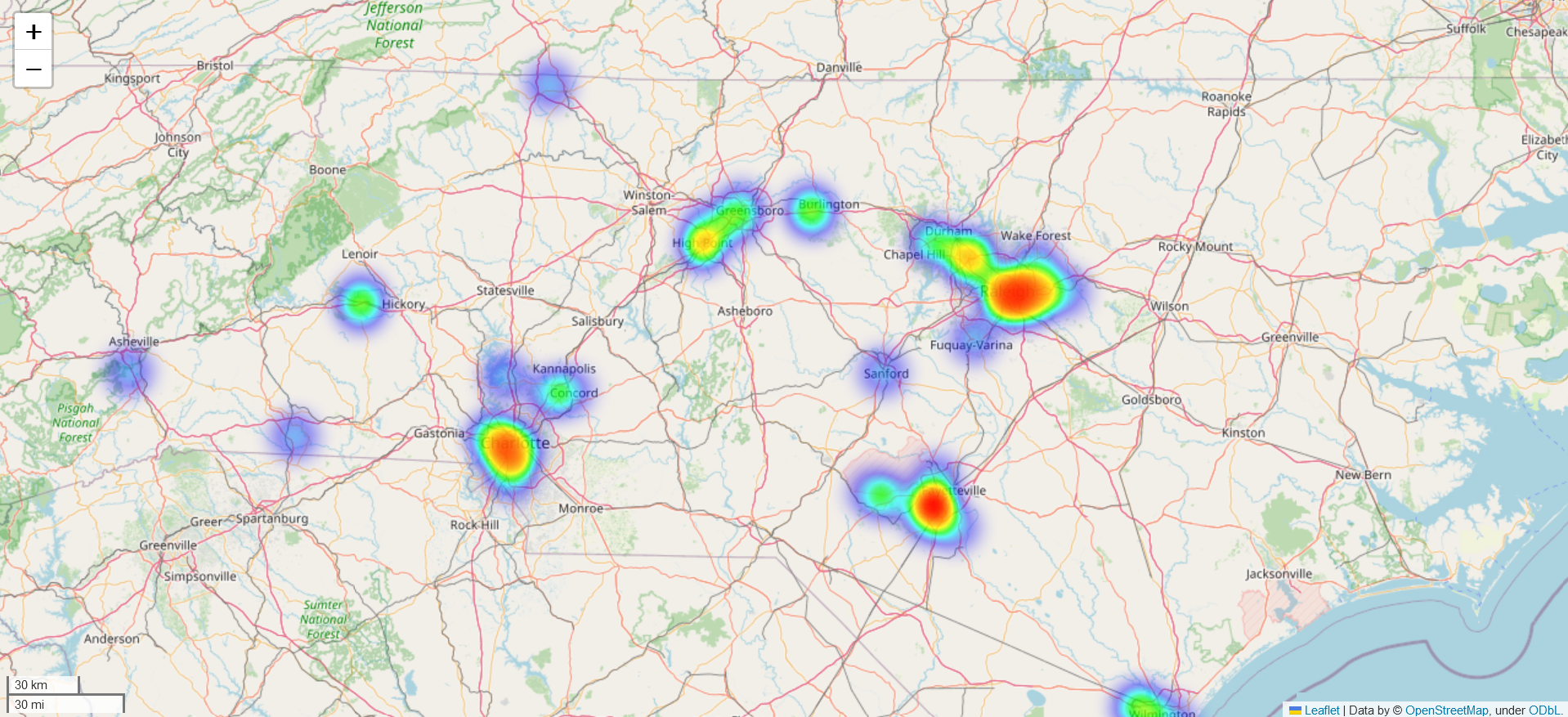}}
\caption{Spatial heat-map weighted by classification confidence of potential crashes being caused by speeding.}
\label{fig:speedroad}
\end{figure}

To demonstrate this implementation, crash and road inventory data from 2018 and 2019 were used to train each road feature model and the road feature data from 2020 were used for visualization. The original HSIS database does not include road name to route identification mappings. Therefore, the roads available for visualization are limited to secondary roads, for which there is a database provided by NCDOT. Each road under consideration will be assessed with respect to its potential to have more crashes with specific contributing factors. The coordinates of each road segment with a predicted contributing factor confidence of over 50 percent were plotted on a spatial heat-map weighted by the confidence percentage, as displayed in Figures \ref{fig:alcroad} - \ref{fig:speedroad}. 

These region-wide risk evaluations demonstrate the potential of ROADFIRST to identify regions of interest at scale for multiple contributing factors, improving the current standard of limiting selections to a specific focus crash and facility type. Specifically, the results from such a critical location identification process can effectively prioritize those locations with more than one contributing factor of interest and help select targeted countermeasures at these locations. Moreover, the practitioners can evaluate the location of interest with ROADFIRST through both crash records and road features, which makes sure the improvements can not only address revealed risk factors but also potential latent causes to future crashes.

\section{CONCLUSIONS}
Despite the effectiveness of the systemic approach to safety and its promotion by the FHWA, the comprehensiveness of the analysis and the efficient use of valuable crash data are constrained due to limitations of the commonly adopted procedure. Specifically, the analysis usually starts by selecting focus crash and facility types, making developed projects less likely to address all aspects of safety concerns at the selected locations. 

In response to such limitations, we developed ROADFIRST, a more comprehensive project development process following the core logic of the systemic approach. Such a process allows users to analyze region-specific risk factors and input all related road features of road segments of interest and identify risks of both potential crash types and potential crash contributing factors for each location. The customization and development of such a procedure using sample data from North Carolina, including validation using data from Minnesota and Maine, has been demonstrated. Traffic-related and driver-related features of each road have been analyzed and their impacts on the sample crash contributing factors, namely alcohol-impaired driving, distracted driving, and speeding, are quantified using the SHAP analysis. An automated risk evaluation model using solely traffic-related features was developed and demonstrated on road segments across North Carolina as well.

Through the proposed procedure, several insights into complex relationships that are only apparent through a process such as this were revealed. These insights generally matched ground truth statistics but provided far more granularity. For example, time of day is a significant influential feature in alcohol-impaired driving model predictions, and the likelihood of classification of a crash as an alcohol-impaired driving crash is heavily positively influenced by crash occurrence between the hours of 11:00 PM and 4:00 AM. Crashes attributed to distracted driving are inversely correlated with speed limit, with possible reasoning being that it is more comfortable to engage in distractions such as cell phones at low speeds. The output of the proposed procedure allows for the derivation of a list of comprehensive contributing factors based on combinations of existing road features, regardless of limitations on crash or facility types. The future direction of this study details the inclusion of a broader contributing factor list and the compound analysis of multiple road features. 

The findings can be further adopted by agencies to develop effective enforcement programs or target countermeasures at specific locations. For example, DUI enforcement should be placed at low traffic density road segments for several hours after 11:00 PM. Such recommendations are better designed by incorporating localized situations. Another direction of future research is to conduct before-after studies to verify the effectiveness of the treatments developed using the theory from this study.  

Despite the innovative nature and verified effectiveness of the proposed system, it should be noted the transferability of the developed system depends on a range of factors, including crash data quality, coverage of road inventory data, crash density, and the significance of the selected contributing factors. In addition, the compound impact of multiple road features on a contributing factor has yet to be specifically investigated. 


%%%%%%%%%%%%%%%%%%%%%%%%%%%%%%%%%%%%%%%%%%%%%%%%%%%%%%%%%%%%%%%%%%%%%%%%%%%%%%%%

%%%%%%%%%%%%%%%%%%%%%%%%%%%%%%%%%%%%%%%%%%%%%%%%%%%%%%%%%%%%%%%%%%%%%%%%%%%%%%%%

%%%%%%%%%%%%%%%%%%%%%%%%%%%%%%%%%%%%%%%%%%%%%%%%%%%%%%%%%%%%%%%%%%%%%%%%%%%%%%%%

%%%%%%%%%%%%%%%%%%%%%%%%%%%%%%%%%%%%%%%%%%%%%%%%%%%%%%%%%%%%%%%%%%%%%%%%%%%%%%%%

\bibliographystyle{ieeetr}
%\bibliography{ARXIV_PAPER}

\end{document}